\documentclass[english]{IEEEtran}
\usepackage[T1]{fontenc}
\usepackage[latin9]{inputenc}
\usepackage{amsmath}
\usepackage{graphicx}



\usepackage{babel}
\begin{document}

\title{Belief Hidden Markov Model for Speech Recognition}
\author{
    \IEEEauthorblockN{{Siwar Jendoubi}\IEEEauthorrefmark{1} \and {Boutheina Ben Yaghlane}\IEEEauthorrefmark{2} \and {Arnaud Martin}\IEEEauthorrefmark{3}}
   \\ \IEEEauthorblockA{\IEEEauthorrefmark{1}{University of Tunis, ISG Tunis, LARODEC Laboratory}
    \\ jendoubi.siouar@laposte.net}
    \\\IEEEauthorblockA{\IEEEauthorrefmark{2}{University of Carthage, IHEC Carthage, LARODEC Laboratory}
    \\ boutheina.yaghlane@ihec.rnu.tn}
    \\\IEEEauthorblockA{\IEEEauthorrefmark{3}{University of Rennes I, IUT de Lannion, UMR 6074 IRISA}
    \\ arnaud.martin@univ-rennes1.fr}
}
\maketitle
\begin{abstract}
Speech Recognition searches to predict the spoken words automatically.
These systems are known to be very expensive because of using several
pre-recorded hours of speech. Hence, building a model that minimizes
the cost of the recognizer will be very interesting. In this paper,
we present a new approach for recognizing speech based on belief HMMs
instead of probabilistic HMMs. Experiments shows that our belief recognizer
is insensitive to the lack of the data and it can be trained using
only one exemplary of each acoustic unit and it gives a good recognition
rates. Consequently, using the belief HMM recognizer can greatly minimize
the cost of these systems. \end{abstract}

\section{Introduction}

The automatic speech recognition is a domain of science that attracts
the attention of the public. Indeed, who never dreamed of talking
with a machine or at least control an apparatus or a computer by voice.
The speech processing includes two major disciplines which are the
speech recognition and the speech synthesis. The automatic speech
recognition allows the machine to understand and process oral information
provided by a human. It uses matching techniques to compare a sound
wave to a set of samples, compounds generally of words or sub-words.
On the other hand, the automatic speech synthesis allows the machine
to reproduce the speech sounds of a given text. Nowadays, most speech
recognition systems are based on the modelling of speech units known
as acoustic unit. Indeed, speech is composed of a sequence of elementary
sounds. These sounds put together make up words. Then, from these
units we seeks to derive a model (one model per unit), which will
be used to recognize continuous speech signal. Hidden Markov Models
(HMM) are very often used to recognize these units. HMM based recognizer
is a widely used technique that allows as to recognize about 80\%
of a given speech signal, but this recognition rate still not yet
satisfying. Also, this method needs many hours of speech for training
which makes the automatic speech recognition task very expensive.

Recently, \cite{Ramasso07a,Ramasso09} extend the Hidden Markov Model
to the theory of belief functions. The belief HMM will avoid disadvantages
of probabilistic HMM which are, generally, due to the use of probability
theory. Belief functions are used in several domains of research where
incertitude and imprecision dominate. They provide many tools for
managing and processing the existent pieces of evidence in order to
extract knowledge and make better decision. They allow experts to
have a more clear vision about their problems, which is helpful for
finding better solutions. What's more, belief functions theories present
a more flexible ways to model uncertainty and imprecise data than
probability functions. Finally, it offers many tools with a higher
ability to combine a great number of pieces of evidence.

Belief HMM gives a better classification rate than the ordinary HMM
when they are applied in a classification problem. Consequently, we
propose to use the belief HMM in the speech recognition process. Finally,
we note that this is the first time where belief functions are used
in speech processing.

In the next section we talk about the probabilistic hidden Markov
model and we define its three famous problems. In Section three we
present the probabilistic HMM recognizer, the acoustic model and the
recognition process. The transferable belief model is introduced in
section four. In section five we will talk about the belief HMM. In
section six, we present our belief HMM recognizer, the belief acoustic
model and the belief recognition process. Finally, experiments are
presented in section seven.

\section{Probabilistic HMM}

A Hidden Markov Model is a combination of two stochastic processes;
the first one is a Markov chain that is characterized by a finite
set%
\footnote{$t$ notes the current instant, it is put in exponent of states for
simplicity.%
} $\Omega_{t}$ of non observable $N$ states (hidden) and the transition
probabilities, $a_{ij}=\textrm{P}\left(s_{j}^{t+1}\mid s_{i}^{t}\right),\,1\leq i,j\leq N$,
between them. The second stochastic process produces the sequence
of $T$ observations which depends on the probability density function
of the observation model defined as $b_{j}\left(O_{t}\right)=\textrm{P}\left(O_{t}\mid s_{j}^{t}\right),\,1\leq j\leq N,\,1\leq t\leq T$
\cite{Rabiner89}, in this paper we use a mixture of Gaussian densities.
The initial state distribution is defined as $\pi_{i}=\textrm{P}\left(s_{i}^{1}\right),\,1\leq i\leq N$.
Hence, an HMM $\lambda\left(A,B,\Pi\right)$ is characterized by the
transition matrix $A=\left\{ a_{ij}\right\} $, the observation model
$B=\left\{ b_{j}\left(O_{t}\right)\right\} $ and the initial state
distribution $\Pi=\left\{ \pi_{i}\right\} $.

There exist three basic problems of HMMs that must be solved in order
to be able to use these models in real world applications. The first
problem is named the evaluation problem, it searches to compute the
probability $P(O/\lambda)$ that the observation sequence $O$ was
generated by the model $\lambda$. This probability can be obtained
using \textit{the forward propagation} \cite{Rabiner89}. Recursively,
it estimates the forward variable: 
\begin{eqnarray}
\alpha_{t}(i) & = & \textrm{P}\left(O_{1}O_{2}\text{\ensuremath{\ldots}}O_{t},q_{t}=s_{i}\mid\lambda\right)\\
\alpha_{t}(i) & = & \left(\sum_{i=1}^{N}\alpha_{t-1}\left(i\right)a_{ij}\right)b_{j}\left(O_{t}\right)
\end{eqnarray}
 for all states and at all time instant. Then, $P(O/\lambda)=\sum_{i=1}^{N}\alpha_{T}\left(i\right)$
is obtained by summing the terminal forward variables. Also, \textit{the
backward propagation} can be used to resolve this problem. Unlike
forward, the backward propagation goes backward. At each instant,
it calculates the backward variable: 
\begin{eqnarray}
\beta_{t}(i) & = & \textrm{P}\left(O_{t+1}O_{t+1}\text{\ensuremath{\ldots}}O_{T}\mid q_{t}=s_{i},\,\lambda\right)\\
\beta_{t}(i) & = & \sum_{j=1}^{N}a_{ij}b_{j}\left(O_{t+1}\right)\beta_{t+1}\left(i\right)
\end{eqnarray}
 finally, $\textrm{P}\left(O\mid\lambda\right)=\sum_{i=1}^{N}\alpha_{t}(i)\beta_{t}(i)$
is obtained by combining the forward and backward variable. The second
problem is named the decoding problem. It searches to predict the
state sequence $S$ that generated $O$. The Viterbi \cite{Rabiner89}
algorithm solves this problem. It starts from the first instant, $t=1$,
for each moment $t$, it calculates $\delta_{t}(i)$ for every state
$i$, then it keeps the state which have the maximum $\delta_{t}=\max_{q_{1},q_{2},\ldots,q_{t-1}}\textrm{P}\left(q_{1},q_{2},\ldots q_{t-1},q_{t}=i,O_{1}O_{2}\ldots O_{t-1}\mid\lambda\right)=\max_{1\leq i\leq N}\left(\delta_{t-1}\left(i\right)a_{ij}\right)b_{j}\left(O_{t}\right)$.
When, the algorithm reaches the last instance $t=T$, it keeps the
state which maximize $\delta_{T}$. Finally, Viterbi algorithm back-track
the sequence of states as the pointer in each moment $t$ indicates.
The last problem is the learning problem, it seeks to adjust the model
parameters in order to maximize $\textrm{P}\left(O\mid\lambda\right)$.
Baum-Welch \cite{Rabiner89} method is widely used. This algorithm
uses the forward and backward variables to re-estimate the model parameters.

\section{Prebabilistic HMM based recognizer}

\subsection{Acoustic model\label{sub:Acoustic-model}}

The acoustic model attempts to mimic the human auditory system, it
is the model used by the HMM-based speech recognizer in order to transform
the speech signal into a sequence of acoustic units, this last will
be transformed into phoneme sequence and finally the desired text
is generated by converting the phoneme sequence into text. Acoustic
models are used by speech segmentation and speech recognition systems.

The acoustic model is composed of a set of HMMs \cite{Rabiner89},
each HMM corresponds to an acoustic unit. To have a good acoustic
model some choices have to be done:

\paragraph{The acoustic unit}

the choice of the acoustic unit is very important, in fact, the number
of them will influence the complexity of the model (more large the
number, more complex the model). If we choose a small unit like the
phone we will have an HMM for every possible phone in the language,
the problem with this choice is that the phone do not model its context.
Such a model is called \textbf{\textit{context independent model}}.
These models are generally used for speech segmentation systems. Other
units that take the context into account can be used as acoustic unit
as the diphone which model the transition between two phones, the
triphone which model the transition between three phones, subwords,
words. These models are called \textbf{\textit{context dependent models}}.
According to \cite{RabinerJuang93}, when the context is greater,
the recognition performance improve.

\paragraph{The model}

for each acoustic unit we associate an HMM, then types of HMM model
and the probability density function of the observation must be chosen.
Generally, left-right models are used for speech recognition and speech
synthesis systems \cite{Rabiner89}. In fact, Speech signal has the
property that it changes over time, then the choice of the left-right
model is justified by the fact that there is no back transitions and
all transitions goes forward. The number of states is fixed in advance
or chosen experimentally. \cite{Carvalho98,Cox98} fixed the number
of state to three. This choice is justified by the fact that most
phoneme acoustic realization is characterized by three sub-segments,
hence we have a state for each sub-segment. \cite{Brugnara93,Toledano03}
used an HMM of six states. Finally, we choose the probability density
function of the observation. They are represented by a mixture of
Gaussian pdf, the number of mixtures is generally chosen experimentally.

The next step, consists on training parameters of each HMM using a
speech corpus that contains many exemplary of each acoustic unit.
Speech segments are transformed into sequence of acoustic vectors
by the mean of a feature extraction method like MFCC, these acoustic
vectors are our sequence of observations.

Then, HMMs are concatenated to each other and we obtain the model
that will be used to recognize the new speech signal. The recognizer
contains three levels; the first one is the \textbf{\textit{syntactic
level}}. It represents all possible word sequences that can be recognized
by our model. The second level is the \textbf{\textit{lexical level}}.
It represents the phonetic transcription (the phoneme sequence) of
each word. Finally, the third level is the acoustic level. It models
the realization of each acoustic unit (in this case the phone).

\subsection{Speech recognition process\label{sub:Speech-recognition-process}}

The model described above is used for the speech recognition process.
Let $S$ be our speech signal to be recognized. Recognizing $S$ consists
on finding the most likely path in the syntactic network. The first
step, is to transform $S$ into a sequence of acoustic vectors using
the same feature extraction method used for training, then we obtain
our sequence of observation $O$. The most likely path is the path
that maximizes the probability of observing $O$ such the model $\textrm{P}\left(O|\lambda\right)$.
This probability can be done either by using the forward algorithm,
or the Viterbi algorithm.

\section{Transferable Belief Model}

The Transferable Belief Model (TBM) \cite{Smets94,Smets00} is a well
used variant of belief functions theories. It is a more general system
than the Bayesian model.

Let $\Omega_{t}=\left\{ \omega_{1},\omega_{2},...,\omega_{n}\right\} $
be our frame of discernment, The agent belief on $\Omega_{t}$ is
represented by the basic belief assignment (BBA) $m^{\Omega_{t}}$
defined from $2^{\Omega}$ to $\left[0,1\right]$. $m^{\Omega_{t}}\left(A\right)$
is the mass value assigned to the proposition $A\subseteq\Omega_{t}$
and it must respect: $\sum_{A\subseteq\Omega_{t}}m^{\Omega_{t}}\left(A\right)=1$.
Also, we can define conditional BBA. Then we can have $m^{\Omega_{t}}$$\left[S^{t-1}\right]\left(A\right)$
which is a BBA defined conditionally to $S^{t-1}\subseteq\Omega_{t-1}$.
If we have $m^{\Omega_{t}}\left(\emptyset\right)>0$, our BBA can
be normalized by dividing the other masses by $1-m^{\Omega_{t}}\left(\emptyset\right)$
then the conflict mass id redistributed and $m^{\Omega_{t}}\left(\emptyset\right)=0$.

Basic belief assignment can be converted into other functions. They
represent the same information under other forms. What's more, they
are in one to one correspondence and they are defined from $2^{\Omega}$
to $\left[0,1\right]$. We will use belief $bel$, plausibility $pl$
and commonality $q$ functions:

\begin{eqnarray}
bel^{\Omega}\left(A\right) & = & \sum_{\emptyset\neq B\subseteq A}m^{\Omega}\left(B\right),\,\forall A\subseteq\emptyset,\, A\neq\emptyset\label{eq:m2bel}\\
m^{\Omega}\left(A\right) & = & \sum_{B\subseteq A}\left(-1\right)^{|A|-|B|}bel^{\Omega}\left(B\right),\,\forall A\subseteq\Omega\label{eq:bel2m}\\
pl^{\Omega}\left(A\right) & = & \sum_{B\cap A=\emptyset}m^{\Omega}\left(B\right),\,\forall A\subseteq\Omega\label{eq:m2pl}\\
m^{\Omega}\left(A\right) & = & \sum_{B\subseteq A}\left(-1\right)^{|A|-|B|-1}pl^{\Omega}\left(\bar{B}\right),\,\forall A\subseteq\Omega\label{eq:pl2m}\\
q^{\Omega}\left(A\right) & = & \sum_{B\supseteq A}m^{\Omega}\left(B\right),\,\forall A\subseteq\Omega\label{eq:m2q}\\
m^{\Omega}\left(A\right) & = & \sum_{A\subseteq B}\left(-1\right)^{|B|-|A|}q^{\Omega}\left(B\right),\,\forall A\subseteq\Omega\label{eq:q2m}
\end{eqnarray}
 Consider two distinct BBA $m_{1}^{\Omega}$ and $m_{2}^{\Omega}$
defined on $\Omega$, we can obtain $m_{1\cap2}^{\Omega}$ through
the TBM conjunctive rule (also called conjunctive rule of combination
CRC) \cite{Smets93} as: 
\begin{equation}
m_{1\cap2}^{\Omega}\left(A\right)=\sum_{B\cap C=A}m_{1}^{\Omega}\left(B\right)m_{2}^{\Omega}\left(C\right),\,\forall A\subseteq\Omega
\end{equation}

Equivalently, we can calculate the CRC via a more simple expression
defined with the commonality function: 
\begin{equation}
q_{1\cap2}^{\Omega}\left(A\right)=q_{1}^{\Omega}\left(A\right)q_{2}^{\Omega}\left(A\right),\,\forall A\subseteq\Omega
\end{equation}

\section{Belief HMM}

Belief HMM is an extension of the probabilistic HMM to belief functions
\cite{Ramasso07a,Ramasso09,Serir11}. Like probabilistic HMM, the
belief HMM is a combination of two stochastic processes. Hence, a
belief HMM is characterized by: 
\begin{itemize}
\item The credal transition matrix $A=\left\{ m_{a}^{\Omega_{t}}\left[S_{i}^{t-1}\right]\left(S_{j}^{t}\right)\right\} $
a set of BBA functions defined conditionally to all possible subsets
of states $S_{i}^{t-1}$, 
\item The observation model $B=\left\{ m_{b}^{\Omega_{t}}\left[O_{t}\right]\left(S_{j}^{t}\right)\right\} $a
set of BBA functions defined conditionally to the set of possible
observation $O_{t}$, 
\item The initial state distribution $\Pi=\left\{ m_{\pi}^{\Omega_{1}}\left(S_{i}^{\Omega_{1}}\right)\right\} $. 
\end{itemize}
The three basic problem of HMM and their solutions are extended to
belief functions. As we know the forward algorithm resolves the evaluation
problem in the probabilistic case. \cite{Ramasso07a} introduced the
\textbf{\textit{credal forward algorithm}} in order to resolve this
problem in the evidential case. It needs as inputs $m_{a}^{\Omega_{t}}\left[S_{i}^{t-1}\right]\left(S_{j}^{t}\right)$
and $m_{b}^{\Omega_{t}}\left[O_{t}\right]\left(S_{j}^{t}\right)$
to calculate the forward commonality: 
\begin{eqnarray}
q_{\alpha}^{\Omega_{t+1}}\left(S_{j}^{t+1}\right) & = & \left(\sum_{S_{i}^{t}\subseteq\Omega_{t}}m_{\alpha}^{\Omega_{t}}\left(S_{i}^{t}\right).q_{a}^{\Omega_{t+1}}[S_{i}^{t}]\left(S_{j}^{t+1}\right)\right)\nonumber \\
 &  & \cap q_{b}^{\Omega_{t+1}}\left[O_{t}\right]\left(S_{j}^{t}+1\right)
\end{eqnarray}
 This last is calculated recursively from $t=1$ to $T$. \cite{Ramasso09}
exploits the conflict of the forward BBA (obtained by using formula
\ref{eq:q2m}) to define an evaluation metric that can be used for
classification to choose the model that best fits the observation
sequence or it can also be used to evaluate the model. Then, given
a model $\lambda$ and an observation sequence of length $T$, the
conflict metric is defined by: 
\begin{align}
L_{c}\left(\lambda\right) & =\frac{1}{T}\sum_{t=1}^{T}\log\left(1-m_{\alpha}^{\Omega_{t+1}}\left[\lambda\right]\left(\emptyset\right)\right)\label{eq:conflict metric 1}\\
\lambda_{*} & =\arg\max_{\lambda}L_{c}\left(\lambda\right)\label{eq:conflict metric2}
\end{align}
 A \textbf{\textit{credal backward algorithm}} is also defined, recursively,
it calculates the backward commonality from $T$ to $t=1$. More details
can be found in \cite{Ramasso07a,Ramasso09}. For the decoding problem,
many solutions are proposed to extend the Viterbi algorithm to the
TBM \cite{Ramasso07a,Ramasso09,Serir11}. All of them search to maximize
the state sequence plausibility. According to the definition given
in \cite{Serir11}, the plausibility of a sequence of singleton states
$S=\left\{ s^{1},s^{2},\ldots,s^{T}\right\} ,\, s^{t}\in\Omega_{t}$
is given by: 
\begin{equation}
pl_{\delta}\left(S\right)=pl_{\pi}\left(s^{1}\right).\prod_{t=2}^{T}pl_{a}^{\Omega_{t}}\left[s^{t-1}\right]\left(s^{t}\right).\prod_{t=1}^{T}pl_{b}\left(s^{t}\right)
\end{equation}

Hence, we can choose the best state sequence by maximizing this plausibility.
For the learning problem, \cite{Ramasso09,Serir11} have proposed
some solutions to estimate model parameters, we will talk about the
method used in this paper. The first step consists on estimating the
mixture of Gaussian models (GMM) parameters using Expectation-Maximization
(EM) algorithm. For each state we estimate one GMM. These models are
used to calculate $m_{b}^{\Omega_{t}}\left[O_{t}\right]\left(S_{j}^{t}\right)$.
\cite{Ramasso09} proposes to estimate the credal transition matrix
independently from the transitions themselves. He uses the observation
BBAs as: 
\begin{eqnarray}
m_{\overline{a}}^{\Omega_{t}\times\Omega_{t+1}} & \propto & \frac{1}{T-1}\label{eq::mbCRCmb}\\
 &  & *\sum_{t=1}^{T}\left(m_{b}^{\Omega_{t}}\left[O_{t}\right]^{\uparrow\Omega_{t}\times\Omega_{t+1}}\cap m_{b}^{\Omega_{t+1}}\left[O_{t+1}\right]^{\uparrow\Omega_{t}\times\Omega_{t+1}}\right)\nonumber 
\end{eqnarray}
 where $m_{b}^{\Omega_{t}}\left[O_{t}\right]^{\uparrow\Omega_{t}\times\Omega_{t+1}}$
and $m_{b}^{\Omega_{t+1}}\left[O_{t+1}\right]^{\uparrow\Omega_{t}\times\Omega_{t+1}}$
are computed using the vacuous extension operator \cite{Smets93}
of the BBA $m_{b}^{\Omega_{t}}\left[O_{t}\right]\left(S_{j}^{t}\right)$
on the cartesian product space as: 
\begin{equation}
m_{b}^{\Omega_{t}\uparrow\Omega_{t}\times\Omega_{t+1}}\left(A\right)=\begin{cases}
m_{b}^{\Omega_{t}}\left(B\right) & \textrm{if}\, A=B\times\Omega_{t+1}\\
0 & \textrm{otherwise}
\end{cases}
\end{equation}

This estimation formula is used by \cite{Serir11} as an initialization
for \textbf{\textit{ITS (Iterative Transition Specialization)}} algorithm.
ITS is an iterative algorithm that uses the credal forward algorithm
to improve the estimation results of the credal transition matrix.
It stops when the conflict metric (formula \ref{eq:conflict metric 1})
converged.

\section{Belief HMM based recognizer}

Our goal is to create a speech recognizer using the belief HMM instead
of the probabilistic HMM. HMM recognizer uses an acoustic model to
recognize the content of the speech signal. Then, we seek to mimic
this model in order to create a belief HMM based one. We should note
that existent parameter estimation methods presented for the belief
HMM cannot be used to estimate model parameters using multiple observation
sequences. This fact should be taken into account when we design our
belief acoustic model.

\subsection{Belief acoustic model\label{sub:Acoustic-model-Belief}}

In the probabilistic case, we use an HMM for each acoustic unit, its
parameters are trained using multiple speech realization of the unit
\cite{RabinerJuang93,Brugnara93,Carvalho98,Toledano03,Cox98}. In
the credal case, a similar model cannot be used. Hence, we present
an alternate method that takes this fact into account.

Let $K$ be the number of the speech realization of a given acoustic
unit. These speech realization are transformed into MFCC feature vectors.
Hence, we obtain $K$ observation sequences. Our training set will
be:$O=\left[O^{1},O^{2},\ldots,O^{K}\right]$ where $O^{k}=\left(O_{1}^{k},O_{2}^{k},\ldots,O_{T_{k}}^{k}\right)$
is the $k^{th}$ observation sequence of length $T_{k}$. These observations
are supposed to be independent to each other. So instead of training
one model for all observation set $O$, we propose to create a belief
model for each observation sequence $O^{k}$. These $K$ models will
be used to represent the given acoustic unit in the recognition process.

Like the acoustic model based on the probabilistic HMM, we have to
make some choices in order to have a good belief acoustic model. In
the first place, we choose the acoustic unit. The same choices of
the probabilistic case can be adopted for the belief case. In the
second place, we choose the model. We should note that we cannot choose
the topology of the belief HMM, this is due to the estimation process
of the credal transition matrix. In other words, the resultant credal
observation model is used to estimate the credal transition matrix
which does not give as the hand to choose the topology of our resultant
model. Consequently, choosing the model in the credal case consists
on choosing the number of states and the number of Gaussian mixtures.
In our case we fix the number of states to three and we choose the
number of Gaussian mixtures experimentally.

\subsection{Speech recognition process}

The belief acoustic model is used in the speech recognition process.
Now, we explain how the resultant model will be used for recognizing
speech signal.

Let $S$ be our speech signal to be recognized. Recognizing $S$ consists
on finding the most likely set of models. The first step, is to transform
$S$ into a sequence of acoustic vectors using the same feature extraction
method used for training, then we obtain our sequence of observation
$O$. This last is used as input for all models. The credal forward
algorithm is then applied, each model gives us an output which is
the value of the conflict metric. An acoustic unit is presented by
a set of models, every model gives a value for the conflict metric.
Then we calculate the arithmetic mean of the resultant values. Finally,
we choose the set of models that optimizes the average of the conflict
metric instead of optimizing the conflict metric, as proposed by \cite{Ramasso09},
using formula \ref{eq:conflict metric2}.

\section{Experiments}

In this section we present experiments in order to validate our approach.
We compare our belief HMM recognizer to a similar one implemented
using the probabilistic HMM.

\begin{figure}
\begin{centering}
\includegraphics[scale=0.6]{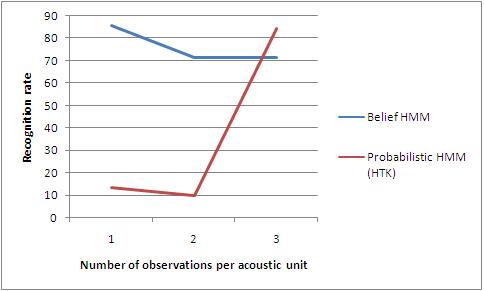} 
\par\end{centering}

\caption{\label{fig:Inf_numb_obser}Influence of the number of observations
on the recognition rate}
\end{figure}

We use MFCC (Mel Frequency Cepstral Coefficient) as feature vectors.
Also, we use a three state HMM and two Gaussian mixtures. Finally,
to evaluate our models we calculate the percent of correctly recognized
acoustic units (number of correctly recognized acoustic unit / total
number of acoustic units). We use a speech corpus that contains speech
realization of seven different acoustic units and we have fifteen
exemplary of each one. Results are shown in figure \ref{fig:Inf_numb_obser}.

The lack of data for training the probabilistic HMM leads to a very
poor learning and the resultant acoustic model cannot be efficient.
Then using a training set that contains only one exemplary of each
acoustic unit leads to have a bad probabilistic recognizer. In this
case our belief HMM based recognizer gives a recognition rate equal
to 85.71\% against 13.79\% for the probabilistic HMM which is trained
using HTK \cite{Young06}. This results shows that the belief HMM
recognizer is insensitive to the lack of data and we can obtain a
good belief acoustic model using only one observation for each unit.
In fact, the belief HMM models knowledge by taking into account doubt,
imprecision and conflict which leads to a discriminative model in
the case of the lack of data.

HTK is a toolkit for HMMs and it is optimized for the HMM speech recognition
process. It is known to be powerful under the condition of having
many exemplary of each acoustic unit. Hence, it needs to use several
hours of speech for training. Having a good speech corpus is very
expensive which influence the cost of the recognition system. Then,
the speech recognition systems are very expensive. Consequently, using
the belief HMM recognizer can greatly minimize the cost of these systems.

\section{Conclusion}

In this paper, we proposed the Belief HMM recognizer. We showed that
incorporating belief functions theory in the speech recognition process
is very beneficial, in fact, it reduces considerably the cost of the
speech recognition system. Future works will be focuced on the case
of the noisy speech signal. Indeed, existent speech recognizer still
not yet good if we have a noisy signal to be decoded.

\bibliographystyle{abbrv}
\bibliography{sans}

\end{document}